\pdfoutput=1

\documentclass[11pt]{article}

\usepackage[]{EMNLP2022}

\usepackage{times}
\usepackage{latexsym}

\usepackage[T1]{fontenc}

\usepackage[utf8]{inputenc}

\usepackage{microtype}

\usepackage{inconsolata}

\usepackage{multirow}
\usepackage{stfloats}
\usepackage{booktabs} 
\usepackage{tipa}
\usepackage{xcolor}

\usepackage{CJK}
\usepackage{algorithm}
\usepackage{algorithmic}
\usepackage{amsmath}
\usepackage{amsfonts}
\usepackage{balance}
\usepackage{graphicx}
\usepackage{subcaption}
\usepackage{epsfig}
\usepackage{url}
\usepackage{array}

\title{Disentangled Text Representation Learning with Information-Theoretic Perspective for Adversarial Robustness}

\author{Jiahao Zhao \qquad Wenji Mao \\
   Institute of Automation, Chinese Academy of Sciences\\
  \texttt{\{zhaojiahao2019,wenji.mao\}@ia.ac.cn} \\}

\begin{document}
\maketitle
\begin{abstract}
Adversarial vulnerability remains a major obstacle to constructing reliable NLP systems. When imperceptible perturbations are added to raw input text, the performance of a deep learning model may drop dramatically under attacks. Recent work argues the adversarial vulnerability of the model is caused by the non-robust features in supervised training. Thus in this paper, we tackle the adversarial robustness challenge from the view of disentangled representation learning, which is able to explicitly disentangle robust and non-robust features in text. Specifically, inspired by the variation of information (VI) in information theory, we derive a disentangled learning objective composed of mutual information to represent both the semantic representativeness of latent embeddings and differentiation of robust and non-robust features. On the basis of this, we design a disentangled learning network to estimate these mutual information. Experiments on text classification and entailment tasks show that our method significantly outperforms the representative methods under adversarial attacks, indicating that discarding non-robust features is critical for improving adversarial robustness.
\end{abstract}

\section{Introduction}

Although deep neural networks have achieved great success in a variety of Natural Language Processing (NLP) tasks, recent studies show their vulnerability to malicious perturbations \cite{goodfellow2015adversarial, jia-liang-2017-adversarial, JiDeepWordBug18, textfooler20}. By adding imperceptible perturbations (e.g. typos or synonym substitutions) to original input text, attackers can generate adversarial examples to deceive the model. Adversarial examples pervasively exist in typical NLP tasks, including text classification \cite{textfooler20}, dependency parsing \cite{zheng-etal-2020-evaluating}, machine translation \cite{zhang-etal-2021-crafting} and many others. These models work well on clean data but are sensitive to imperceptible perturbations. Recent studies indicate that they are likely to rely on superficial cues rather than deeper, more difficult language phenomena, and thus tend to make incomprehensible mistakes under adversarial examples \cite{jia-liang-2017-adversarial, branco-etal-2021-shortcutted}.

Tremendous efforts have been made to improve the adversarial robustness of NLP models. Among them, the most effective strategy is adversarial training \cite{Li_Qiu_2021,wang2021infobert,dong2021towards}, which minimizes the maximal adversarial loss. As for the discrete nature of text, another effective strategy is adversarial data augmentation \cite{min-etal-2020-syntactic, zheng-etal-2020-evaluating, ivgi-berant-2021-online}, which augments the training set with adversarial examples to re-train the model. Guided by the information of perturbation space, these two strategies utilize textual features as a whole to make the model learn a smooth parameter landscape, so that it is more stable and robust to adversarial perturbations.

As adversarial examples pervasively exist, previous research has studied the underlying reason for this \cite{goodfellow2015adversarial, nips2016noise, nips2018moredata, tsipras2018robustness, Ilyas2019bugs}. One popular argument \cite{Ilyas2019bugs} is that adversarial vulnerability is caused by the non-robust features. While classifiers strive to maximize accuracy in standard supervised training, they tend to capture any predictive correlation in the training data and may learn predictive yet brittle features, leading to the occurrence of adversarial examples. These non-robust features leave space for attackers to intentionally manipulate them and trick the model. Therefore, discarding the non-robust features can potentially facilitate model robustness against adversarial attacks, and this issue has not been explored by previous research on adversarial robustness in text domain.

To address the above issue, we take the approach of disentangled representation learning (DRL), which decomposes different factors into separate latent spaces. In addition, to measure the dependency between two random variables for disentanglement, we take an information-theoretic perspective with the Variation of Information (VI). Our work is particularly inspired by the work of \citet{cheng-etal-2020-info-theoretic}, which takes an information-theoretic approach to text generation and text-style transfer. As our focus is on disentangling robust and non-robust features for adversarial robustness, our work is fundamentally different from the related work in model structure and learning objective design.

In this paper, we tackle the adversarial robustness challenge and propose an information-theoretic Disentangled Text Representation Learning (DTRL) method. Guided with the VI in information theory, our method first derives a disentangled learning objective that maximizes the mutual information between robust/non-robust features and input data to ensure the semantic representativeness of latent embeddings, and meanwhile minimizes the mutual information between robust and non-robust features to achieve disentanglement. On this basis, we leverage adversarial data augmentation and design a disentangled learning network which realizes task classifier, domain classifier and discriminator to approximate the above mutual information.  Experimental results show that our DTRL method improves model robustness by a large margin over the comparative methods.

The contributions of our work are as follows:

\begin{itemize}
\item {We propose a disentangled text representation learning method, which takes an information-theoretic perspective to explicitly disentangle robust and non-robust features for tackling adversarial robustness challenge.}
\item {Our method deduces a disentangled learning objective for effective textual feature decomposition, and constructs a disentangled learning network to approximate the mutual information in the derived learning objective.}
\item {Experiments on text classification and entailment tasks demonstrate the superiority of our method against other representative methods, suggesting eliminating non-robust features is critical for adversarial robustness.}
\end{itemize}

\section{Related work}
\paragraph{Textual Adversarial Defense}
To defend adversarial attacks, empirical and certified methods have been proposed. Empirical methods are dominant which mainly include adversarial training and data augmentation. Adversarial training \cite{Miyato17MKI,Li_Qiu_2021,wang2021infobert,dong2021towards, li2021textdefender} regularizes the model with adversarial gradient back-propagating to the embedding layer. Adversarial data augmentation \cite{min-etal-2020-syntactic, zheng-etal-2020-evaluating, ivgi-berant-2021-online} generates adversarial examples and retrains the model to enhance robustness. 
Certified robustness \cite{jia-etal-2019-certified, huang-etal-2019-ibp, Shi2020transformer} minimizes an upper bound loss of the worst-case examples to guarantee model robustness.  Besides, adversarial example detection \cite{zhou-etal-2019-disp,mozes-etal-2021-frequency,bao-etal-2021-defending} identifies adversarial examples and recovers the perturbations. Unlike these previous methods, we enhance model robustness from the view of DRL to eliminate non-robust features. 
 
\paragraph{Disentangled Representation Learning}
Disentangled representation learning (DRL) encodes different factors into separate latent spaces, each with different semantic meanings. The DRL-based methods are proposed mainly for image-related tasks. \citet{Pan_Niu_Zhang_Zhang_2021} propose a general disentangled learning method based on information bottleneck principle \cite{tishby2000information}. Recent work also extends DRL to text generation tasks, e.g. style-controlled text generation \cite{Mixpoet-20, cheng-etal-2020-info-theoretic}. Different from the DRL-based text generation work that uses encoder-decoder framework to disentangle style and content in text, our work develops the learning objective and network structure to disentangle robust and non-robust features for adversarial robustness.

Existing DRL-based methods for adversarial robustness have solely applied in image domain \cite{yang2021classdisentanglement, Yang_Guo_Wang_Xu_2021, kim2021distilling}, mainly based on the VAE. Different from continuous small perturbation pixels in image that are suitable for generative models, text perturbations are discrete in nature, which are hard to deal with using generative models due to their overwhelming training costs. With adversarial data augmentation, our method uses a lightweight layer with cross-entropy loss for effective disentangled representation learning.

\section{Preliminary}

The Variation of Information (VI) is a fundamental metric in information theory that quantifies the independence between two random variables. Given two random variables $U$ and $V$, $\mathrm{VI}(U ; V)$ is defined as:
\begin{equation} \label{eq:vi}
\mathrm{VI}(U ; V)=\mathrm{H}(U)+\mathrm{H}(V)-2 \mathrm{I}(U ; V),
\end{equation}
where $\mathrm{H}(U)$ and $\mathrm{H}(V)$ are the Shannon entropy, and $\mathrm{I}(U ; V) = \mathbb{E}_{p(u, v)}\left[\log \frac{p(u, v)}{p(u) p(v)}\right]$ is the mutual information between $U$ and $V$. 

The VI is a positive, symmetric metric. It obeys the triangle inequality \cite{kraskov2005hierarchical}, that is, for any random variables $U$, $V$ and $W$:
\begin{equation} \label{eq:tri}
\mathrm{VI}(U ; V) + \mathrm{VI}(U ; W) \ge \mathrm{VI}(V ; W).
\end{equation}
Equality occurs if and only if the information of $U$ is totally divided into that of $V$ and $W$.

\section{Problem Definition}
Given a victim model $f_{v}$ and an original input $x \in X$ where $X$ is input text set, an attack method $\mathcal{A}$ is applied to search perturbations to construct an adversarial example $ \hat{x} \in \hat{X}$ which fools the model prediction (i.e. $f_{v}(x)\ne f_{v}(\hat{x})$ ). Adversarial attacks can be regarded as data augmentation. For random variables $X, Y \sim p_{\mathcal{D}}(x, y)$ where $Y$ is the set of class labels, $(x,y)$ is the observed value, $D$ is a dataset and $p_{\mathcal{D}}$ is the data distribution. The goal of adversarial robustness is to build a classifier $f(y|x)$ that is robust against adversarial attacks.

\begin{figure*}[h] 
\centering
\includegraphics{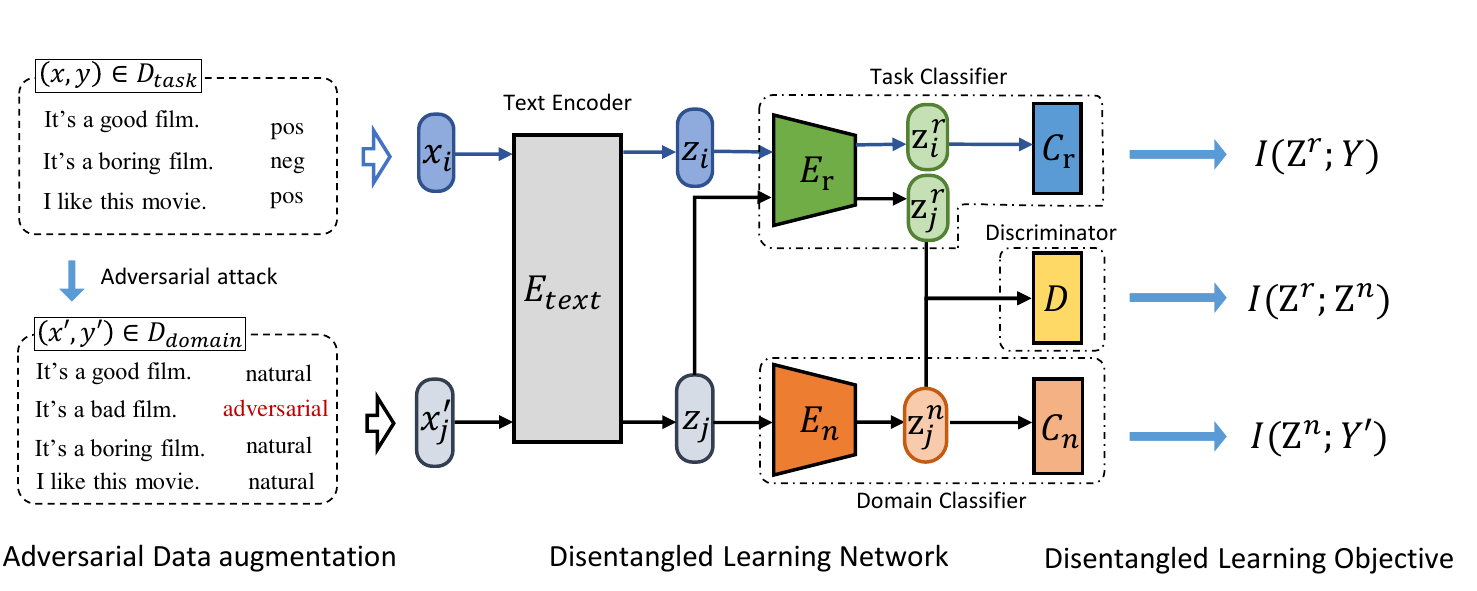} 
\caption{Overview of our proposed DTRL method for adversarial robustness.}
\label{fig:model}
\vspace{-1em}
\end{figure*}

\section{Proposed Method}
The overall architecture of our proposed method is shown in Fig.\ref{fig:model}. We first apply adversarial attacks to augment the original textual data. We then design the disentangled learning objective to separate features into robust and non-robust ones. Finally, we construct the disentangled learning network to implement the learning objective. 
\subsection{Adversarial Data Augmentation}
As adversarial examples have different patterns other than clean data like word frequency \cite{mozes-etal-2021-frequency} and fluency \cite{lei2022phrase}, we use adversarial examples to guide the non-robust features learning. To efficiently disentangle robust and non-robust features, we employ adversarial data augmentation to get adversarial examples for the extention of training set. 

We denote original training set as $D_{task} = \{x_i,y_i\}^N_{i=1}$, where $x$ is input text, $y$ is task label (e.g. \emph{positive} or \emph{negative}), $x \in X$ and $y \in Y$.  We apply adversarial data augmentation to $D_{task}$ and get adversarial examples $\hat{x} \in \hat{X}$. We then construct domain dataset $D_{domain}=\{x^\prime_j,y^\prime_j\}^M_{j=1}$, where $x^\prime$ is input text or adversarial example, $y^\prime$ is domain label (e.g. \emph{natural} or \emph{adversarial}), $x^\prime \in \{X, \hat{X}\}$, $y^\prime \in Y^\prime$ and $Y^\prime$ is the set of domain labels.

\subsection{Disentangled Learning Objective}
We propose our learning objective that disentangles the robust and non-robust features, and buide the approximation method to estimate mutual information in the derived learning objective. We use the VI in information theory to measure the dependency between latent variables for disentanglement. In contrast to the computational alternative of generative model like variational autoencoder (VAE), our method considers the discrete nature of text and develops an effective VI-guided disentangled learning  technique with less computational cost.

\subsubsection{Learning Objective Derivation}
We start from $\mathrm{VI}(Z^r;Z^n)$ to measure the independence between robust features $Z^r$ and non-robust features $Z^n$. By applying the triangle inequality of VI (Eq.(\ref{eq:tri})) to $X$, $Z^r$ and $Z^n$, we have
\begin{equation} \label{eq:3}
\mathrm{VI}(X;Z^r)+\mathrm{VI}(X;Z^n) \ge \mathrm{VI}(Z^r;Z^n),
\end{equation}
where the difference between $\mathrm{VI}(X;Z^r)+\mathrm{VI}(X;Z^n)$ and $\mathrm{VI}(Z^r;Z^n)$ represents the degree of disentanglement. By simplifing Eq.(\ref{eq:3}) with the definition of VI (Eq.\ref{eq:vi}), we have
\begin{equation} \label{eq:sub}
\begin{aligned}
& \mathrm{VI}(X;Z^r)+\mathrm{VI}(X;Z^n)-\mathrm{VI}(Z^r;Z^n) \\
=& 2 \mathrm{H}(X)+2[\mathrm{I}(Z^r;Z^n)-\mathrm{I}(X; Z^r)-\mathrm{I}(X;Z^n)].
\end{aligned}
\end{equation}

Then for a given dataset, $\mathrm{H}(X)$ is a constant positive value. By dropping $\mathrm{H}(X)$ and the coefficient from Eq.(\ref{eq:sub}), we have
\begin{equation} \label{eq:simply}
\begin{aligned}
 \mathrm{VI}&(X;Z^r)+\mathrm{VI}(X;Z^n)-\mathrm{VI}(Z^r;Z^n) \\
>& \mathrm{I}(Z^r;Z^n)-\mathrm{I}(X; Z^r)-\mathrm{I}(X;Z^n).
\end{aligned}
\end{equation}

As in Eq.(\ref{eq:simply}), the robust and non-robust features are symmetrical and interchangeable, we further differentiate them by introducing supervised information. Recent study shows that without inductive biases, it is theoretically impossible to learn disentangled representations \cite{locatello2019challenging}. Therefore, we leverage the task label in $Y$ and domain label in $Y^\prime$ to supervise robust and non-robust feature learning respectively. 

Specifically, encoding $X$ into $Z^r$ to predict output $Y$ forms a Markov chain $ X \rightarrow Z^r \rightarrow Y$ and obeys the data processing inequality (DPI):
\begin{equation} \label{eq:zr}
\mathrm{H}(X) \ge \mathrm{I}(X;Z^r) \ge \mathrm{I}(Z^r;Y),
\end{equation}
where $\mathrm{I}(\cdot; \cdot)$ is the mutual information between two latent variables. Similarly, we have Markov chain $X \rightarrow Z^n \rightarrow Y^\prime$ with the DPI formulism:
\begin{equation}  \label{eq:zn}
\mathrm{H}(X) \ge \mathrm{I}(X;Z^n) \ge \mathrm{I}(Z^n;Y^\prime).
\end{equation}

Finally, by combining Eq.(\ref{eq:zr}) and Eq.(\ref{eq:zn}) to Eq.(\ref{eq:simply}), we get the upper bound of our disentangled learning objective:
\begin{equation} \label{eq:dis}
\begin{aligned}
& \mathrm{I}(Z^r;Z^n)-\mathrm{I}(Z^r;Y)-\mathrm{I}(Z^n;Y^\prime) \ge \\
 & \mathrm{I}(Z^r;Z^n)-\mathrm{I}(X; Z^r)-\mathrm{I}(X;Z^n).
\end{aligned}
\end{equation}


\subsubsection{Mutual Information Approximation}
As the mutual information in Eq.(\ref{eq:dis}) is computationally intractable, it is difficult to calculate it directly. Thus we adopt the variational Bayesian method to approximate the estimation of mutual information. 

To maximize $\mathrm{I}(Z^r;Y)$ and $\mathrm{I}(Z^n;Y^\prime)$, we derive their variational lower bounds\footnote{In this work, $X,Y,Y^\prime,Z^r,Z^n$ are random variables, and $x,y,y^\prime,z^r,z^n$ are instances of these random variables.}. For $\mathrm{I}(Z^r;Y)$:
\vspace{-0.3em}
\begin{equation}
\begin{aligned}
 I(Z^r ; Y) &=\mathbb{E}_{p(z^r, y)} \log p(z^r \mid y)-\mathbb{E}_{p(y)} \log p(y) \\ 
& \ge \mathbb{E}_{p(z^r, y)} \log q(z^r \mid y)+H(Y),
\vspace{-0.3em}
\end{aligned}
\end{equation}
where $p(z^r, y)$ is the data distribution, and $q(z^r|y)$ is the variational posterior of robust features $Z^r$ conditioned on task label in $Y$. Similarly, the variational lower bound for $I(Z^n ; Y^\prime)$:
\vspace{-0.3em}
\begin{equation}
\begin{aligned} I(Z^n ; Y^\prime) &=\mathbb{E}_{p(z^n, y^\prime)} \log p(z^n \! \mid \! y^\prime) \! - \! \mathbb{E}_{p(y^\prime)} \log p(y^\prime) \\ 
& \ge \mathbb{E}_{p(z^n, y^\prime)} \log q(z^n \! \mid \! y^\prime)+H(Y^\prime),
\end{aligned}
\end{equation}
where $p(z^n, y^\prime)$ is the data distribution, and $q(z^n|y^\prime)$ is the variational posterior of non-robust features $Z^n$ conditioned on domain label in $Y^\prime$.

We then estimate $\mathrm{I}(Z^r;Z^n)$ with the \textit{density-ratio-trick} \cite{pmlr-v80-kim18b, Pan_Niu_Zhang_Zhang_2021}. By the definition of mutual information, we have
\vspace{-0.3em}
\begin{equation}
\mathrm{I}(Z^r;Z^n) = \mathbb{E}_{p(z^r, z^n)}\left[\log \frac{p(z^r,z^n)}{p(z^r)p(z^n)}\right],
\end{equation}
where $\frac{p(z^r,z^n)}{p(z^r)p(z^n)}$ is the ratio between joint distribution $p(z^r,z^n)$ and the product of the marginal distribution $p(z^r)p(z^n)$. We further train a discriminator $D$ to estimate whether it is sampled from $p(z^r,z^n)$ or $p(z^r)p(z^n)$, and minimize $\mathrm{I}(Z^r;Z^n)$ in an adversarial training manner:
\vspace{-0.3em}
\begin{equation}
\begin{aligned}
\min\max& [\mathbb{E}_{p(z^r) p(z^n)} \log D(z^r, z^n)\\
&+ \mathbb{E}_{p(z^r, z^n)} \log (1-D(z^r, z^n))].
\end{aligned}
\end{equation}


\vspace{-0.7em}
\subsection{Disentangled Learning Network}

Based on the above disentangled learning objective, we propose the disentangled learning network consisting of four components. Text encoder first converts text to vector representation. The other three components then estimate the mutual information in the disentangled learning objective. The robust features are captured by task classifier, while the non-robust ones are captured by domain classifier. Discriminator estimates and minimizes the mutual information between robust and non-robust features for disentanglement.

\vspace{-0.5em}
\subsubsection{Text Encoder}

Given examples $(x_i,y_i) \in D_{task}$ and $(x^\prime_j,y^\prime_j) \in D_{domain}$, we use large-scale pre-trained model to encode input texts:
\begin{equation}
z_i = E_{\mathrm{text}}(x_i),
z_j = E_{\mathrm{text}}(x^\prime_j),
\end{equation} 
where $E_{\mathrm{text}}()$ is the pre-trained model (e.g. BERT), and $z_i$ and $z_j$ are vector representations. We assume that $z$ can be disentangled into robust features $z^r$ and non-robust features $z^n$.
\begin{algorithm}[t]
\caption{Disentangled representation learning}
\begin{algorithmic}[1]
\label{alg:main}
\REQUIRE  Task data $\mathcal{D}_{\mathrm{task}} \in \{x, y\}$, domain data $\mathcal{D}_{\mathrm{dom}} \in \{x^\prime, y^\prime\}$, parameters $\theta$ of encoders and classify layers $\{ E_{\mathrm{text}},E_r, C_r, E_n,C_n \}$, parameters $\phi$ of discriminator $D$
\WHILE {not converge}
	\STATE Sample batch data $\{x_i, y_i\}_{i\in B}$ from $\mathcal{D}_{\mathrm{task}}$
	\STATE $z^r_i = E_r(E_{\mathrm{text}}(x_i))$
	\STATE $\mathcal{L}_{\mathrm{task}}= \frac{1}{B} \sum_{i=1}^{B} \mathcal{L}_{\mathrm{CE}}(C_r(z^r_i),y_i)$
	\STATE Sample batch data $\{x_j^\prime, y_j^\prime\}_{j\in B}$ from $\mathcal{D}_{\mathrm{dom}}$
	\STATE $z^r_j = E_r(E_{\mathrm{text}}(x_j^\prime)), z^n_j = E_n(E_{\mathrm{text}}(x_j^\prime))$
	\STATE $\mathcal{L}_{\mathrm{dom}}= \frac{1}{B} \sum_{j=1}^{B} \mathcal{L}_{\mathrm{CE}}(C_n(z^n_j),y_j^\prime)$
	\STATE $\! \mathcal{L}_{\theta} \! = \! \mathcal{L}_{\mathrm{task}} \! + \mathcal{L}_{\mathrm{dom}} \! - \! \frac{1}{B} \! \sum_{i=1}^{B} \! \mathrm{log}D([z^r_j;z^n_j])$
	\STATE Update $\theta \leftarrow \nabla_{\theta} \mathcal{L}_{\theta}$
	\STATE Resampling on batch axis to indices $k\in B$ 
	\STATE $\mathcal{L}_{\phi}= \frac{1}{B} \sum_{j=1}^{B}$
	\STATE $\quad - \mathrm{log} (1-D([z^r_j;z^n_j])) - \mathrm{log} D([z^r_k;z^n_k])$
	\STATE Update $\phi \leftarrow \nabla_{\phi} \mathcal{L}_{\phi}$
\ENDWHILE
\end{algorithmic}
\end{algorithm}

\vspace{-0.5em}

\subsubsection{Task Classifier to Estimate $I(Z^r;Y)$}
Mutual information $I(Z^r ;Y)$ is used to measure the dependency between robust features $Z^r$ and task label $Y$. We use $I(Z^r ;Y)$ to guide encoder $E_r$ to learn robust features: 
\begin{equation}
z_i^r = E_r(z_i),
z_j^r = E_r(z_j),
\end{equation} 
where $z_i^r$ and $z_j^r$ are the corresponding robust features, and $E_r$ captures the robust features $Z^r$ by learning to predict the class label in  $Y$:
\begin{equation}
\mathcal{L}_{\mathrm{task}} = \mathcal{L}_{\mathrm{CE}}(C_r(z^r_i),y_i),
\end{equation}
where $C_r$ is a task classification layer and $\mathcal{L}_{\mathrm{CE}}$ is cross-entropy loss. $E_r$ and $C_r$ are implemented with multi-layer perceptron (MLP).
\subsubsection{Domain Classifier to Estimate $I(Z^n;Y^\prime)$}
Encoder $E_n$ captures the non-robust features $Z^n$ by learning to predict the domain label in $Y^\prime$:
\begin{equation}
\begin{aligned}
z^n_j &= E_n(z_j), \\
\mathcal{L}_{\mathrm{domain}} &= \mathcal{L}_{\mathrm{CE}}(C_n(z^n_j),y_j^\prime),
\end{aligned}
\end{equation}
where $z^n_j$ are non-robust features and $C_n$ is domain classification layer. $E_n$ and $C_n$ are implemented with MLP.

\subsubsection{Discriminator to Estimate $I(Z^r;Z^n)$}
To disentangle robust and non-robust features, we use a discriminator $D$ to estimate the $I(Z^r;Z^n)$:
\begin{equation}
L_{disen} = - \mathrm{log} (1-D([z^r_j;z^n_j])) - \mathrm{log} D([z^r_k;z^n_k]),
\end{equation}
where $[\cdot;\cdot]$ is concatenation operation, $(z^r_j, z^n_j)$ is sampled from the joint distribution $p(z^r,z^n)$, and $(z^r_k,z^n_k)$ is sampled from the product of marginal distribution $p(z^r)p(z^n)$ by shuffling samples aligned with the batch axis \cite{pmlr-v80-mine}. $D$ is implemented with MLP.  

\subsection{Optimization}
Our disentangled learning network is optimized by adversarial training in an end-to-end manner. We use $\theta$ for the parameters of $\{ E_{\mathrm{text}},E_r, C_r, E_n,C_n \}$  and $\phi$ for the parameters of discriminator $D$. The overall loss is:
\begin{equation}
\min_{\theta}(\mathcal{L}_{task}+\mathcal{L}_{domain}+\max_{\phi}\mathcal{L}_{disen}).
\end{equation}

We use the reparameterization trick \cite{Kingma2014} to approximate the gradients of $E_r$ and $E_n$. During the training process, $\theta$ and $\phi$ are updated alternately (see Algorithm \ref{alg:main} for more details). 
We pre-train two classifiers and discriminator for a few iterations before the adversarial training stage to ensure their initial learning ability.

\section{Experiments}
In this section, we first evaluate the effectiveness of our method for adversarial robustness. We then give a detailed analysis on model performance.

\subsection{Datasets and Tasks}

We choose two typical tasks for evaluating adversarial defense methods, text classification and textual entailment. We conduct experiments on three benchmark datasets. Movie Reviews (MR) \cite{pang2005MR} and SST-2 \cite{socher-etal-2013-recursive} are sentiment classification datasets, with each sentence labeled into \{positive, negative\}. Stanford Natural Language Inference (SNLI) is a textual entailment dataset \cite{bowman2015SNLI}, with each pair of sentences labeled into \{entailment, neutral, contrast\}. The statistics are given in Table\ref{tab:dataset}.


\begin{table}[h]
    \small
  \centering
  \begin{tabular}{cccccc}
    \toprule
    Dataset & Train & Test & Classes & Avg. Length \\
    \midrule
    MR & 9K & 1067 & 2 & 20 \\
	SST-2 & 67K & 872 & 2 & 8 \\
    SNLI & 550K & 10K & 3 & 11 \\
  \bottomrule
\end{tabular}
  \caption{Statistics of three datasets.}
  \label{tab:dataset}
\vspace{-1.5em}
\end{table}

\subsection{Victim Model and Attack Methods}
Following the convention of evaluating adversarial robustness, we take BERT \cite{devlin-etal-2019-bert} as the victim model which is fine-tuned on each task with the whole training set. We use two recent representative attack methods.

\textbf{Deepwordbug} \cite{JiDeepWordBug18} is a character-level attack. The perturbation space is character insertion/deletion/swap/substitution and restricted with edit distance to maintain the original meaning.

\textbf{Textfooler} \cite{textfooler20} is a word-level attack. The perturbation space is synonym substitutions. The words close to the original word in counter-fitted word embedding are considered a synonym set. Substitutions are checked with part-of-speech and semantic similarity.

Following the practice of prior work \cite{alzantot-etal-2018-generating,textfooler20} in evaluating adversarial robustness, we use the same 1,000 randomly selected examples from the test set for MR and SNLI, and the whole test set for SST-2. 
Adversarial data augmentation is applied to the training sets with the above two attack methods. For all attack methods, we use the toolkit TextAttack\footnote{\url{https://github.com/QData/TextAttack}} \cite{morris2020textattack} with the default setting (e.g. query limit, similarity constraint and synonym set). 

\subsection{Comparative Methods} %
We compare our proposed DTRL with four advanced adversarial training and data augmentation methods, a general DRL method and an improved BERT fine-tuning method. For fair comparison, all the  methods use BERT as text encoder.

\textbf{VIBERT} \cite{mahabadi2021variational} is a information theoretic method to implement information bottleneck principle that functions as suppress irrelevant features for improved BERT fine-tuning. 

\textbf{ADA} is the standard adversarial data augmentation method that trains the model using the mixture of normal and adversarial data.

\textbf{ASCC} \cite{dong2021towards} is an adversarial sparse convex combination method that estimates the word substitution attack space with convex hull and uses it as a regularization term.
\textbf{TA-VAT} \cite{Li_Qiu_2021} is a token-aware virtual adversarial training method that uses a token-level normalization ball to constrain the perturbation.
\textbf{InfoBERT} \cite{wang2021infobert} is a adversarial training method with BERT fine-tuning using two regularizers. Information bottleneck regularizer suppresses noise features and anchored feature regularizer increases the dependence between local and global features.

\textbf{DisenIB}  \cite{Pan_Niu_Zhang_Zhang_2021} is a supervised disentangled learning method that implements information bottleneck principle. DisenIB uses CNN to reconstruct input images. Transferring to text domain, we replace CNN with an RNN decoder and change the reconstruction loss to language modeling loss for self-reconstruction. 
\vspace{-0.5em}
\subsection{Implementation Details}
For all comparative methods, \emph{bert-base-uncased} \cite{devlin-etal-2019-bert} is used as text encoder. We use the last hidden layer output of token [CLS] as sentence embedding. 
 We run our experiments on one Tesla V100 GPU with 32 GB memory. Details of hyper-parameters are provided in Appendix \ref{sec:appendix}. After parameter searching, we report the best results of all methods. Our codes, trained models and augmented data are available at \url{https://anonymous.4open.science/r/DTRL}. 



\vspace{-0.5em}
\subsection{Experimental Results}
\begin{table*}[ht]
\setlength\extrarowheight{2pt}
	\small
    \centering
    \begin{tabular}{clcccccccc}
    \toprule
    \textbf{Datasets} & \textbf{Score} & \textbf{BERT} & \textbf{VIBERT} & \textbf{ADA} & \textbf{ASCC} & \textbf{TA-VAT} & \textbf{InfoBERT} & \textbf{DisenIB} & \textbf{DTRL}\\
    \midrule
	\multirow{3}{*}{\textbf{MR}} & 
	 Clean         & 86.5 & 88.0 & 82.8 & 84.9 & \textbf{89.4} & 88.4 & 85.0 & 87.0 \\
     & Deepwordbug & 15.3 & 13.6 & 16.6 & 16.0 & 19.9 & 17.9 & 20.9 & \textbf{35.0} \\
     & Textfooler  &  4.4 &  5.0 &  7.1 &  4.0 &  7.2 &  6.1 &  8.5 & \textbf{19.3} \\
    \midrule
	\multirow{3}{*}{\textbf{SST-2}} & 
	 Clean         & 92.4 & 92.7 & \textbf{93.5} & 88.0 & 92.8 & 93.0 & 90.2 & 92.2 \\
     & Deepwordbug & 17.7 & 19.8 & 24.4 & 23.2 & 20.5 & 19.2 & 29.8 & \textbf{40.8} \\
     & Textfooler  &  4.4 &  5.6 &  8.3 &  5.3 &  7.8 &  9.9 &  8.7 & \textbf{17.8}\\
    \midrule
	\multirow{3}{*}{\textbf{SNLI}} & 
	 Clean         & 89.1 & 89.1 & 89.6 & 82.7 & 88.7 & \textbf{90.2} & 86.7 & 89.7 \\
     & Deepwordbug &  6.6 &  8.8 & 17.0 & 12.4 & 12.9 &  7.9 & 13.9 & \textbf{24.1} \\
     & Textfooler  &  1.7 &  2.2 &  5.1 &  3.9 &  3.6 &  2.9 &  7.3 & \textbf{14.6} \\
    \bottomrule
    \end{tabular}
    \caption{Robustness comparison of different methods. All methods are built on BERT, with the best performing scores marked in \textbf{bold}. Clean refers to the accuracy of clean data, and Deepwordbug and Textfooler refer to the accuracies under corresponding attacks.}
    \label{tab:multi}
\vspace{-1.5em}
\end{table*}

In our experiments, we first report the main experimental results of our method and the comparative methods. We then illustrate the latent space to visualize the quality of disentanglement. We also evaluate the transferability of robustness and sensitivity of mutual information estimation.
\subsubsection{Results on Model Robustness}

\begin{figure*}[h] 
\centering
  \includegraphics[scale=0.369]{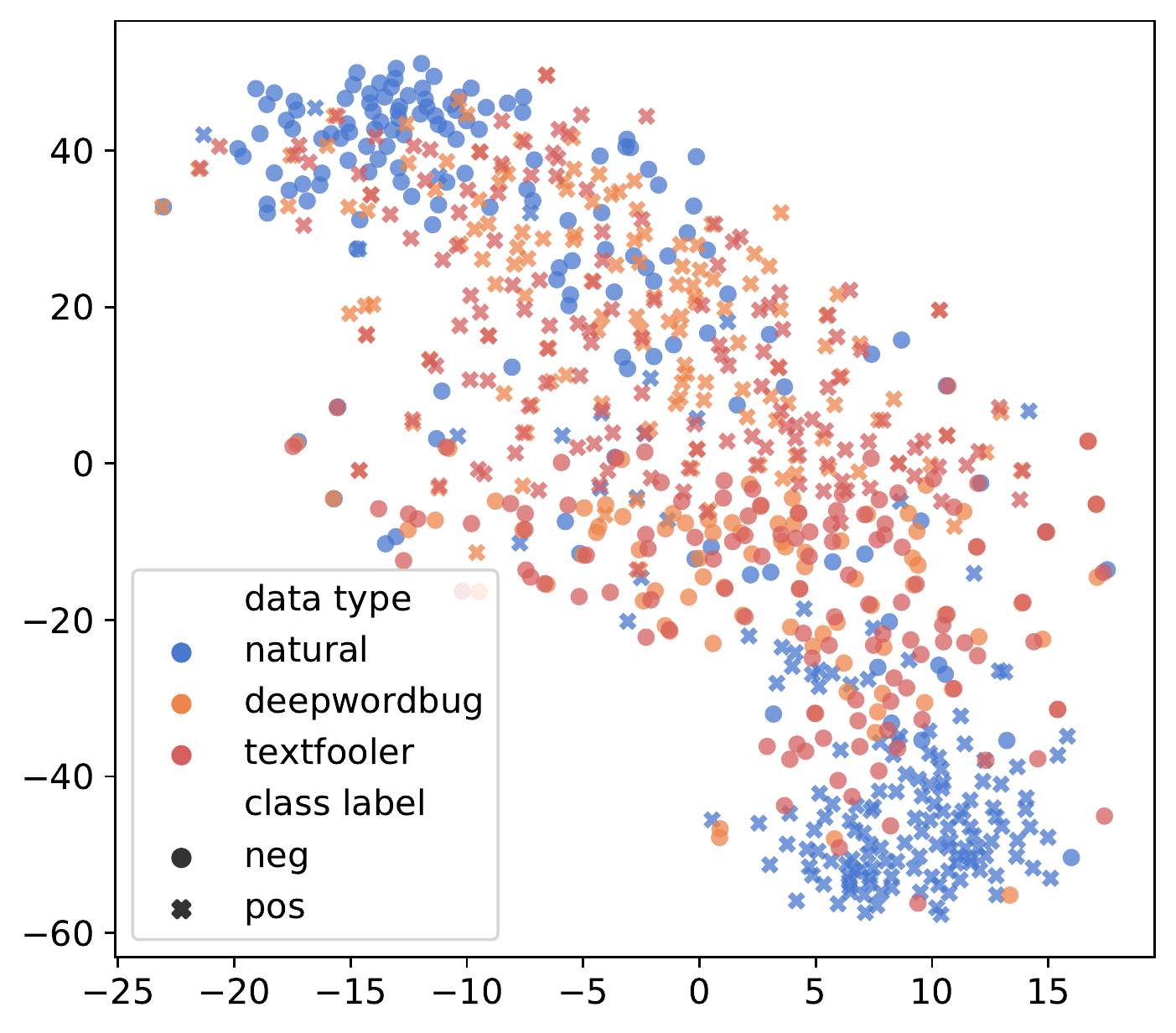}
  \includegraphics[scale=0.369]{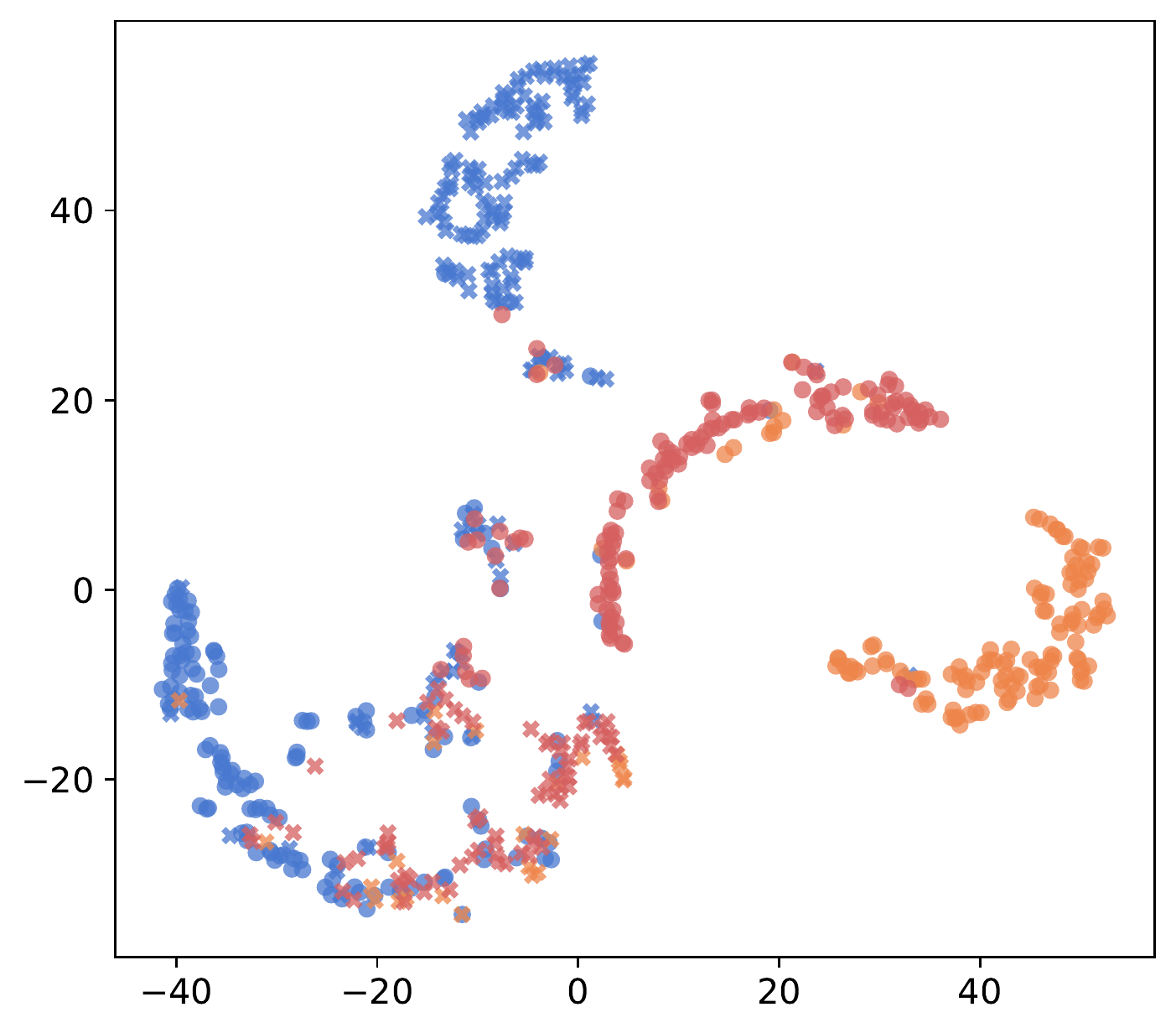}
  \includegraphics[scale=0.369]{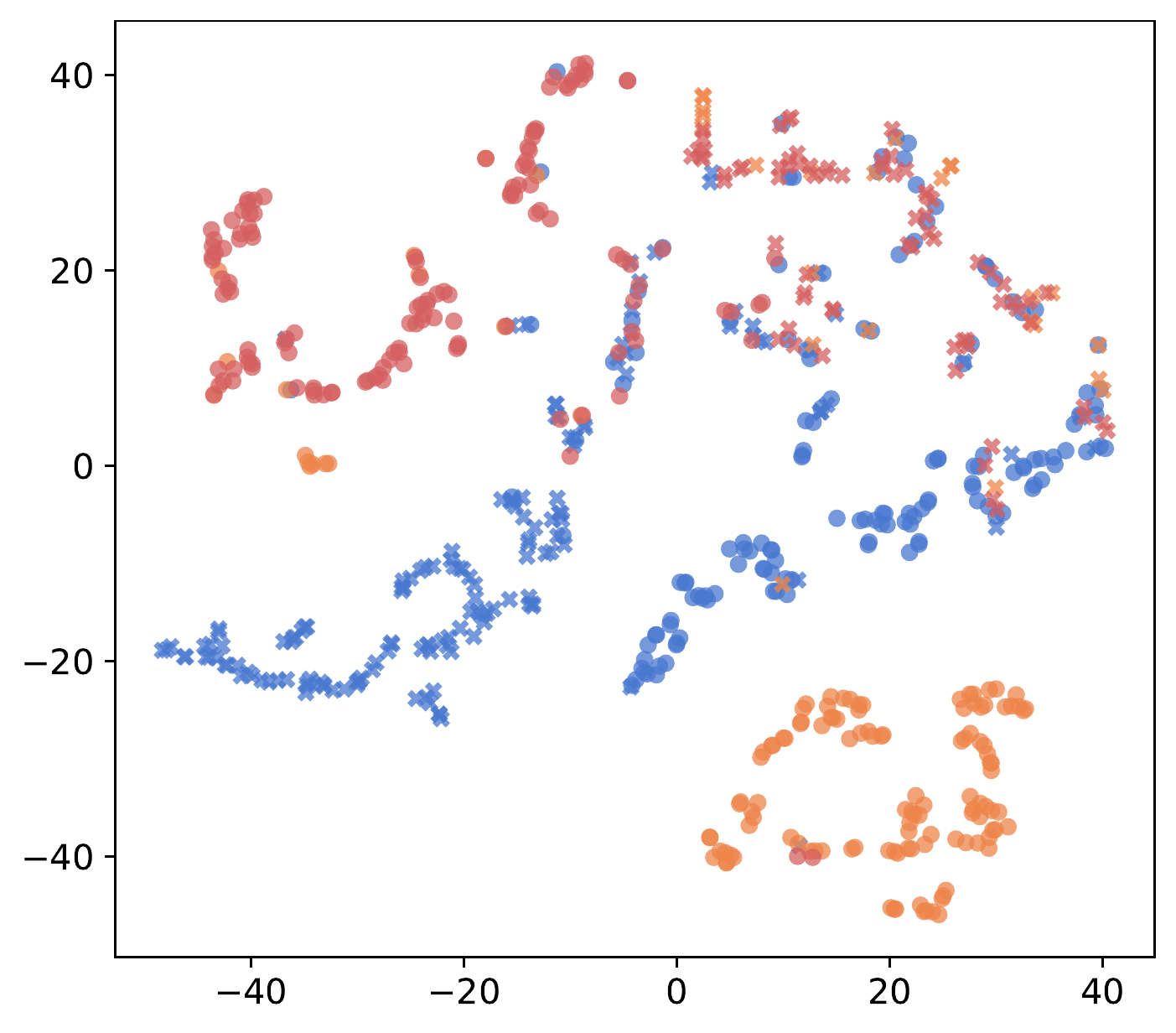}
\caption{Latent space visualization with t-SNE. Left figure illustrates  BERT embeddings, and middle and right figures for robust/non-robust features. Domain labels are marked with colors and task labels with dot and cross.}
\label{fig:embedding}
\vspace{-1.5em} 
\end{figure*}

Table \ref{tab:multi} reports the comparative results of different methods on the accuracies of clean data and under attacks. For clean data, the performances of our DTRL are comparable to those of the original BERT. One possible reason for this is that our disentangled learning objective can  restrain the model from relying on spurious cues, and thus facilitate the exploration of more difficult and robust language phenomena. Under attacks, our DTRL achieves the highest accuracy values and outperforms all other methods by a large margin. 

Comparing the results of ADA and DTRL, it shows the advantages of disentangled learning, as both methods need adversarial data augmentation at the training stage. Under attacks, DTRL increases model robustness by a large margin compared to ADA. The results indicate that DTRL is more effective to leverage adversarial examples to improve robustness. We also report empirical results of robustness transferability in Section \ref{sec:transfer}.

Compared with information bottleneck (IB) guided methods VIBERT and InfoBERT, we can see that InfoBERT achieves better accuracy values on clean data, while DTRL significantly improves the model robustness under attacks. The IB principle learns compressed representations which removing irrelevant features, leading to better performances on clean data. Contrastly, our DTRL explicitly disentangles robust and non-robust features and achieves better results under attacks than IB guided methods. We further visualize the disentangled latent embeddings in Section \ref{sec:visualize}.

The DRL-based DisenIB and our DTRL both achieve better results for adversarial robustness than those of other methods. DisenIB uses the original examples to learn irrelevant features with reconstruction loss while our DTRL directly learns non-robust features with adversarial data augmentation. DTRL outperforms DisenIB in adversarial robustness, indicating that adversarial examples contains more non-robust features than the original examples do. Also, the loss of text decoder in DisenIB behaves quite differently than classification layer, resulting in the difficulty of balancing task loss and disentanglement loss. 

In summary, the above results verify the effectiveness of our method. Due to space limit, the results of ablation study are given in Appendix \ref{sec:ablation}.

\subsubsection{Visualization of Latent Space} \label{sec:visualize}

To visulize the latent space, we first randomly sample 300 examples from the test set of MR, then perform attacks against BERT and our DTRL to get adversarial examples respectively. We compress the representations of BERT embedding and DTRL robust and non-robust features into 2D by t-SNE \cite{la08tsne}. 

Figure \ref{fig:embedding} visualizes the disentangled embeddings in our model. The left figure shows the last layer hidden state of BERT. The middle and right figures show the robust and non-robust features respectively. The domain of examples is colored in blue (natural example), red (Textfooler) and orange (Deepwordbug). The truth label of examples is marked with dot and cross. From the right figure, it can be seen that initially, adversarial examples are located between positive and negative examples. 

We also notice different preferences between robust and non-robust features. From middle figure of robust features, examples are distinct with class labels and grouped with domain (i.e. natural or adversarial). While in the left figure of non-robust features, examples are better grouped by colors which stand for the domain of examples. This indicates non-robust features are mainly guided by domain labels, while robust features mainly focus on  task labels. To some degree, robust and non-robust feature clustering tends to overlap, and better mutual information estimation can potential facilitate the improvement of robustness.

\subsubsection{Transferability of Robustness} \label{sec:transfer}

\begin{table}[ht]
	\small
\setlength\extrarowheight{2pt}
    \centering
    \begin{tabular}{clcccc}
    \toprule
	\textbf{Datasets} & \textbf{Model} & \textbf{textbugger} & \textbf{iga} & \textbf{pso} & \textbf{pwws} \\
    \midrule
	\multirow{2}{*}{\textbf{MR}} & 
	BERT  & 24.4 &  8.6 & 5.8 &  9.3 \\
    & DTRL & 26.6 & 27.5 & 7.7 & 24.6 \\
    \midrule
	\multirow{2}{*}{\textbf{SST-2}} & 
	BERT  & 28.2 & 12.1 & 7.9 & 12.3\\
    & DTRL & 30.0 & 13.9 & 9.7 & 20.3\\
    \midrule
	\multirow{2}{*}{\textbf{SNLI}} & 
	BERT  & 2.5 & 0.8 & 6.3 & 1.3 \\
    & DTRL & 7.8 & 5.5 & 7.9 & 6.2 \\
    \bottomrule
    \end{tabular}
    \caption{Accuracy of BERT and DTRL under attacks.}
\vspace{-1em}
    \label{tab:transfer}
\end{table}

To explore the robustness transferability of our DTRL, we evaluate model robustness with four extra attacks and report model accuracy under attack in Table \ref{tab:transfer}. We select attack methods with different perturbation spaces: iga\footnote{Improved genetic algorithm based word substitution method from \citet{wang-2019-iga}} uses counter-fitted word embedding to search synonym substitutions, while pso \cite{zang-2020-pso} and pwws \cite{ren-2019-pwws} use HowNet and WordNet respectively. textbugger \cite{li-2019-textbugger} consists of character-level attack and word substitution attack. 

Compared to BERT, DTRL consistently improves model robustness under four attacks. This indicates that disentangled robust features can improve robustness across different perturbation spaces. Although this empirical finding does not guarantee the robustness improvement under other attack methods, it sheds light on future research to enhance model robustness intrinsically.

\subsubsection{Effect of Mutual Information Estimation} \label{sec:ablation}

We further investigate the sensitivity of mutual information (MI) estimation, as its estimation is the key to disentanglement.

\paragraph{Batch Size}
\begin{figure}[h] 
\centering
  \includegraphics[scale=0.55]{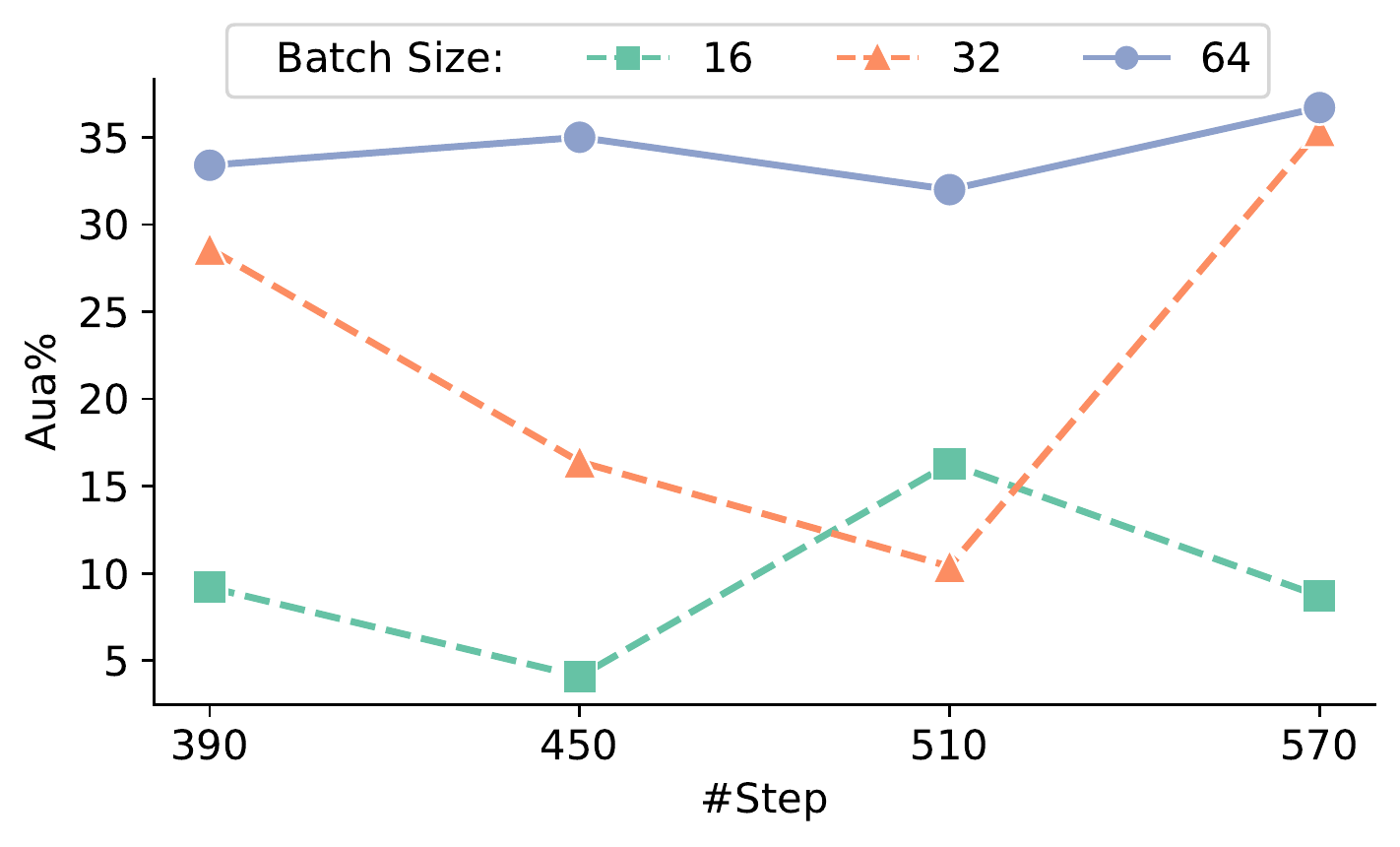}
\caption{Robustness results of MR under Deepwordbug attack. Aua\% is  accuracy under attack and \#Step is training step.}
\vspace{-1em}
\label{fig:line} 
\end{figure}

\begin{figure}[h]
\centering
  \includegraphics[scale=0.55]{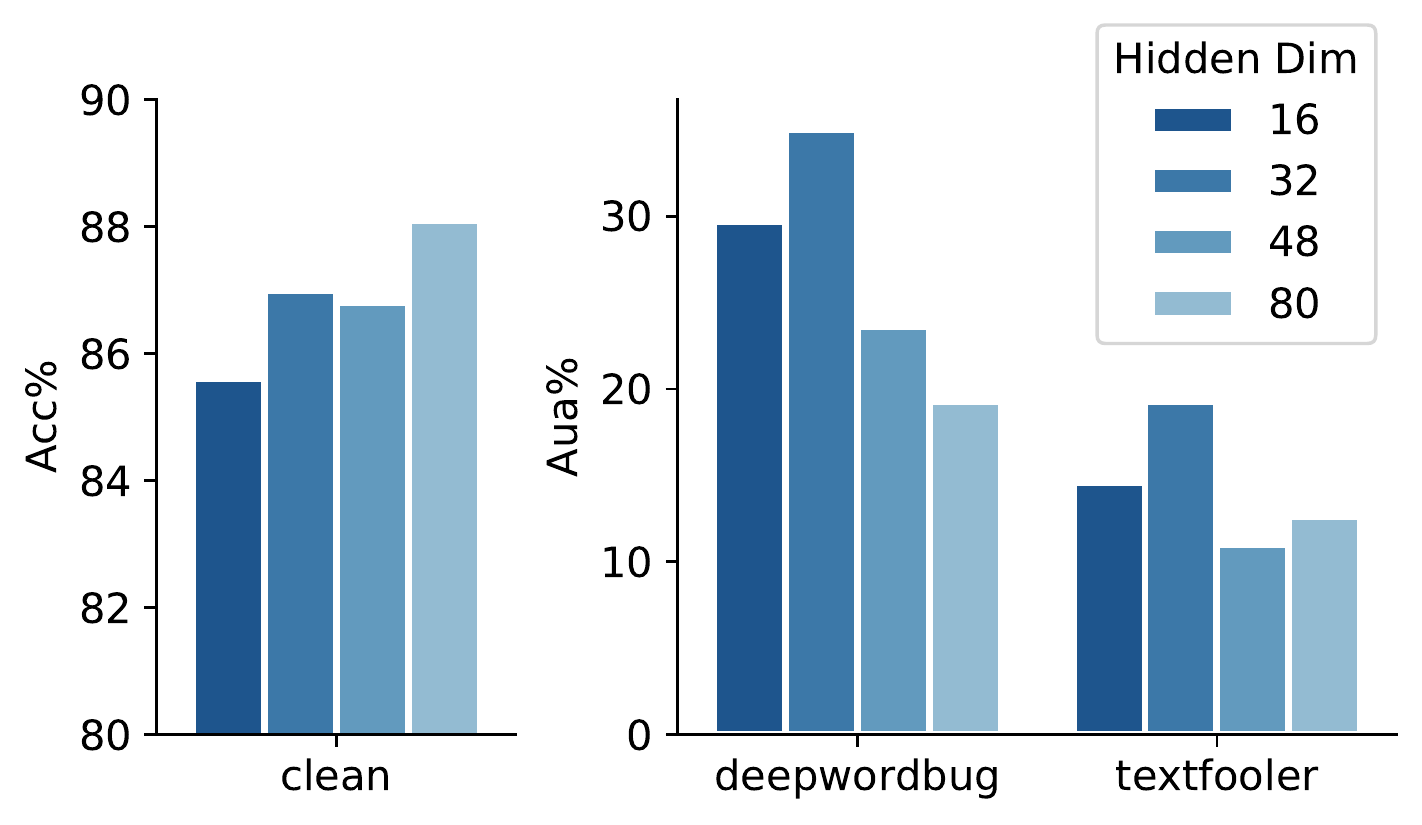}
\caption{Sensitivity of dimension. Acc\% is the accuracy of clean data. Aua\% is the accuracy under attack.}
\vspace{-1.0em}
 \label{fig:bar} 
\end{figure}
We estimate MI using the density-ratio-trick \cite{pmlr-v80-kim18b, Pan_Niu_Zhang_Zhang_2021} which shuffles the samples along the batch axis. Thus we examine the effect of batch size. Figure \ref{fig:line} is the accuracy of DTRL under deepwordbug attack on MR dataset. We can observe that small batch size makes the robustness low and unstable, while large batch leads to consistently better robustness.

\paragraph{Hidden Dimension}

Previous work show that estimating MI is hard, especially when the dimension of the latent embedding is large \cite{pmlr-v80-mine,pmlr-v97-poole19a,pmlr-v119-club}. On the other hand, a large hidden dimension can improve the representative ability of model. We examine this trade-off by altering the hidden dimension of robust and non-robust embedding layers. Figure \ref{fig:bar} shows the results of the accuracies of clean data and under attack. We observe that a larger hidden dimension can increase model accuracy of clean data. On the contrary, when the dimension is too large, it will degrade model robustness.

\vspace{-0.5em}
\section{Conclusion}
\vspace{-0.5em}
In this paper, we propose the disentangled text representation learning method DTRL for adversarial robustness guided with information theory. Our method derives the disentangled learning objective and constructs the disentangled learning network which learns the disentangled representations.

\section*{Limitations}
The disentangled learning objective in our method is derived from the VI with several approximations, which makes it a loose bound, and consequently its mutual information estimation lacks a tight bound either. Another limitation in our method is that the adversarial data augmentation we use for non-robust feature learning is a relatively time-consuming offline method compared to self-reconstruction, albeit it is more effective than self-reconstruction as shown in the comparative results of Table \ref{tab:multi} between our DTRL and DisenIB method.


\section*{Ethical Considerations}
Our research studies disentangled text representation learning to enhance model robustness under adversarial attacks. The potential risk is that the way we disentangle non-robust features may reversely inform the design of adversarial attack methods to increase the vulnerabilities of machine learning models.

\bibliography{anthology,articleegc}
\bibliographystyle{acl_natbib}

\newpage
\appendix
\section{Additional Implementation Details}
\label{sec:appendix}
\paragraph{Baseline Implementation}
The BERT baseline has 110M parameters with 12 layer transformers and hidden size of 768. For methods TA-VAT, ASCC and InfoBERT, we use the implementation of textdefender \footnote{\url{https://github.com/RockyLzy/TextDefender}} \cite{li2021textdefender}. 
For VIBERT method, we use the implementation of \citet{mahabadi2021variational}. \footnote{\url{https://github.com/rabeehk/vibert}}.
 We implement DisenIB based on \citet{Pan_Niu_Zhang_Zhang_2021}. \footnote{\url{https://github.com/PanZiqiAI/disentangled-information-bottleneck}}
For all methods in experiment, the training epoch is 10, 10 and 5 for MR, SST-2 and SNLI respectively. For VIB, we consider $\beta$ of $\{10^{-5},10^{-4},10^{-3},10^{-2}\}$. For ASCC, we consider weight of regularization $\alpha$ of $\{10,3,2,1\}$ and weight of of KL distance of $\{1,4,5\}$. For DisenIB, we consider the weight of estimation of $\{0.5,1, 1.5,2.0,2.5\}$ and the weight of reconstruction of $\{0.5,1, 1.5,2.0,2.5\}$. We consider batch size of $\{16,32,64,96\}$ and learning rate of $\{1e^{-5},1e^{-4},5e^{-5}\}$. We use early stopping and select the best accuracy model on test set. 
\paragraph{Hyper-Parameters} We use the same setting in different datasets. For DTRL, the architecture parameters of $E_r$ and $E_n$ are the same. Both are 3 layer MLP and shapes are:[768, 768], [768,384], [384, 32]. $C_r$ and $C_n$ are one layer MLP shape as [32, \#label type]. $D$ take the concatenate of $z^r$ and $z^n$ as input, output the mutual information between them. $D$ is a 3 layer MLP and shapes are: [64, 128], [128, 256], [256,1]. For the architecture of compared methods, we basically use the default setting of their implementation. Details can be found in our code. 

Table \ref{tab:hyper} lists the hyperparameter configurations for best-performing DTRL model on three datasets.
 
\paragraph{Visualization Parameters} We visualize hidden embeddings using t-SNE. For the hyper-parameters of t-SNE, we consider the iteration of \{300, 500, 1000, 2000, 3000, 5000, 6000\} and perplexity of \{10, 20, 30, 50, 100, 150, 200, 250, 300\}. We draw the cherry picking figures in Fig.\ref{sec:visualize}.

\section{Adversarial Data Augmentation}
\paragraph{Victim Model} is a well trained model as the attack target to generate adversarial examples. For MR dataset, we finetune BERT on MR training set ourselves. For SST-2 and SNLI dataset, we use the \emph{bert-base-uncased-snli} and \emph{bert-base-uncased-sst2} provided by TextAttack Model Zoo \footnote{\url{https://textattack.readthedocs.io/en/latest/3recipes/models.html}}.
\paragraph{Dataset}
We use three English and balanced datasets: MR, SST-2 and SNLI. We use the MR dataset provided by TextFooler \footnote{\url{https://github.com/jind11/TextFooler}} which used 90\% of data as training set and 10\% as the test set. We use the GLUE version of \emph{SST-2} dataset \footnote{\url{https://huggingface.co/datasets/glue/viewer/sst2}}. We use \emph{snli 1.0} provided by Stanford Natural Language Processing Group \footnote{\url{https://nlp.stanford.edu/projects/snli/}}. We augment adversarial examples using the training set of these datasets. We use the full training sets of MR and SST-2. Due to the time limitation, we use half of the training sets of SNLI.

\paragraph{Attack Methods} 
We use the implementation of TextAttack with the default setting. We notice that different work may adjust the constraint parameters of attack methods in different ways. For consistent comparison, we stick to the default parameter setting in this paper. 

Adversarial data augmentation is time-consuming, so we release all augmented data used in experiments for ease of replication. The statistic of augmented data is listed in Table \ref{tab:augsta}. 

\begin{table}[hb]
  \small
  \centering
  \begin{tabular}{lcccc}
    \toprule
      & \multicolumn{2}{c}{\textbf{Deepwordbug}} & \multicolumn{2}{c}{\textbf{Textfooler}} \\
	\cmidrule(lr){2-3} \cmidrule(lr){4-5}  
	\textbf{Datasets} & \textbf{\#A.E.} & \textbf{Avg. P.W.}\% & \textbf{\#A.E.}& \textbf{Avg. P.W.}\% \\
    \midrule
    \textbf{MR}    & 7878 & 19.37 & 9065 & 19.37 \\
    \textbf{SST-2} & 51501 & 34.54 & 59072 & 31.89 \\
    \textbf{SNLI}  & 248246 & 20.21 & 257282 & 7.66 \\
  \bottomrule

\end{tabular}
  \caption{Statistics of augmented data of MR, SST-2 and SNLI datasets. Deepwordbug and Textfooler are the attack methods used for augmentation respectively. \textbf{\#A.E.} represents the number of adversarial examples and \textbf{Avg P.W.} \% represents average word perturbation rate. }
  \label{tab:augsta}
\end{table}
\paragraph{Text Encoder Comparison}
\begin{table*}[ht]
\setlength\extrarowheight{2pt}
	\small
    \centering
    \begin{tabular}{lccccccccc}
    \toprule
    \multirow{2}{*}{Encoder} & \multicolumn{3}{c}{\textbf{MR}}  & \multicolumn{3}{c}{\textbf{SST-2}} & \multicolumn{3}{c}{\textbf{SNLI}}\\
    \cmidrule(lr){2-4}\cmidrule(lr){5-7} \cmidrule(lr){8-10} 
     & Clean  & Deepwordbug & Textfooler & Clean  & Deepwordbug & Textfooler & Clean  & Deepwordbug & Textfooler \\
    \midrule
    BERT       & 86.5 & 15.3 & 4.4 & 92.4 & 17.7 & 4.4 & 89.1 & 6.6 & 1.7 \\
	\midrule
    RoBERTa    & 95.6 & 18.9 & 5.2 & 94.0 & 17.0 & 4.7 & 91.2 & 2.9 & 3.4 \\
	DistilBERT & 86.4 & 17.1 & 3.9 & 90.0 & 14.2 & 2.7 & 87.4 & 1.5 & 1.5 \\
	ALBERT     & 89.7 & 15.0 & 3.4 & 92.6 & 13.6 & 3.9 & 89.1 & 1.0 & 1.3 \\
  \bottomrule
    \end{tabular}
    \caption{Performance comparison of different methods on MR, SST-2 and SNLI. Clean stands for the accuracy of clean data, and  Deepwordbug and Textfooler refer to the accuracy under corresponding attacks.}
    \label{tab:encoder}
\end{table*}

Table \ref{tab:encoder} shows the performance of different encoder. Most of the models are provided by TextAttack Model Zoo. Specifically, there are \emph{roberta-base-mr}, \emph{roberta-base-sst2}, \emph{distilbert-base-uncased-mr}, \emph{distilbert-base-cased-sst2}, \emph{distilbert-base-cased-snli}, \emph{albert-base-v2-mr}, \emph{albert-base-v2-sst2}, \emph{albert-base-v2-snli}. Lastly, we finetune roberta-base on snli with batch size 64, learning rate 2e-5 for 3 epoch.

In terms of performance on clean data, RoBERTa outperforms BERT, while DistilBERT and ALBERT are close to BERT. In terms of accuracy under attacks, the performances of the four encoders are close to each other. Thus we choose BERT as text encoder in our experiments for simplification (note that VIBERT, ASCC, TA-VAT and InfoBERT are original built on BERT in their published papers).

\begin{table*}[ht]
\setlength\extrarowheight{2pt}
	\small
  \centering
  \begin{tabular}{llll}
    \toprule
    \textbf{Hyperparameters} & \textbf{MR} & \textbf{SST-2} & \textbf{SNLI} \\
    \midrule
    Layers of encoder $E_r$ & 3 & 3 & 3 \\
    Layers of encoder $E_n$ & 3 & 3 & 3 \\
    Dimension of encoder $E_r$ & 768,384,32 & 768,384,24 & 768,384,32 \\
	Dimension of encoder $E_n$ & 768,384,32 & 768,384,24 & 768,384,32 \\
	Layers of $C_r$ & 1 & 1 & 1 \\
	Layers of $C_n$ & 1 & 1 & 1 \\
	Dimension of $C_r$ & 32 & 24 & 32 \\
	Dimension of $C_n$ & 32 & 24 & 32 \\
	Layers of $D$ & 3 & 3 & 3 \\
	Dimension of $D$ & 128,256,1 & 128,256,1 & 128,256,1 \\
	Batch size & 64 & 64 & 64 \\
	Warmup steps & 300 & 200 & 2000 \\
	Total training steps & 600 & 400 & 3000 \\
	Learning rate & 5e-5 & 5e-5 & 5e-5 \\
	Optimizer of $D$ & Adam & Adam & Adam \\
	Optimizer of others & AdamW & AdamW & AdamW \\
  \bottomrule
\end{tabular}
  \caption{Hyperparameters for the best-performing DTRL model on dataset MR, SST-2 and SNLI.}
  \label{tab:hyper}
\end{table*}



\section{Ablation Study} \label{sec:ablation}
We conduct an ablation study to test the design of our DTRL. Our DTRL is an integrated method. Without adversarial data augmentation (ADA), domain classifier (DC) and discriminator (D), DTRL degenerates to BERT fine-tuning. Without disentangled learning  domain classifier (DC) and discriminator (D), our DTRL degenerates to ADA. Replacing domain classifier with self-reconstruction decoder (AE), DTRL is similar to DisenIB. We borrow the results from Table \ref{tab:multi} and report the ablation results in Table \ref{tab:ablation}. Without disentangled learning (-DC, -D) and using self-reconstruction for disentanglement learning (-D, +AE), the model performance for adversarial robustness is poor, and our full model using adversarial examples for disentangled representation learning benefits adversarial robustness the most. 

\begin{table}[ht]
  \small
  \centering
  \begin{tabular}{lccc}
    \toprule
    \textbf{Model} & \textbf{Clean}  & \textbf{DWB} & \textbf{TF} \\
    \midrule
    DTRL \hspace{3em} (full model) & 92.2 & 40.8 & 17.8\\
    \midrule
    -ADA, -D, +AE \hspace{0em} (DisenIB)   & 90.2 & 29.8 &  8.7\\
    -D, -DC       \hspace{4.3em} (ADA)   & 93.5 & 24.4 &  8.3\\
    -ADA, -D, -DC \hspace{1em} (BERT)             & 92.4 & 17.7 &  4.4\\
  \bottomrule
\end{tabular}
  \caption{Ablation study on SST-2 dataset. Clean, DWB and TF stand for the accuracies of clean data, under Deepwordbug and  Textfooler attacks respectively.}
  \label{tab:ablation}
\end{table}

\section{Case Analysis}
\begin{figure*}
     \centering
     \begin{subfigure}[b]{\textwidth}
         \centering
         \includegraphics[width=\textwidth]{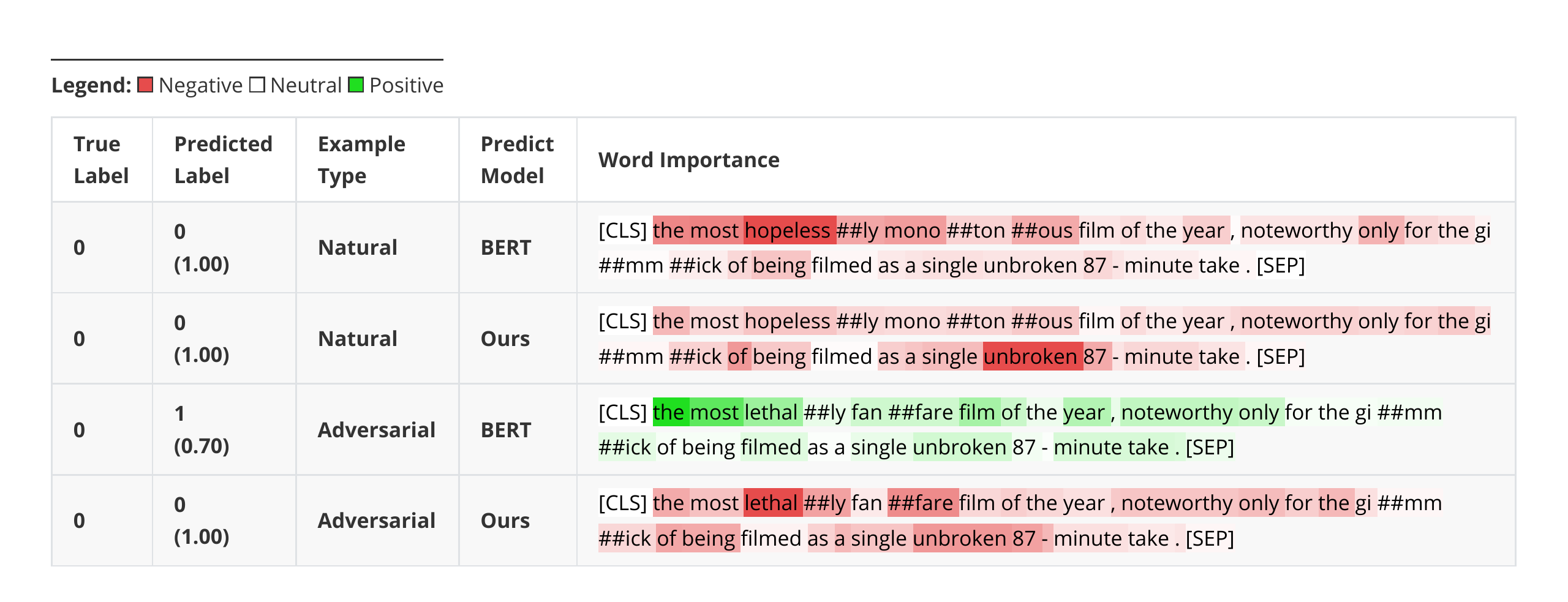}
         \caption{Input text: \emph{the most hopelessly monotonous film of the year , noteworthy only for the gimmick of being filmed as a single unbroken 87-minute take .}}
         \label{fig:case1}
     \end{subfigure}
     \hfill
     \begin{subfigure}[b]{\textwidth}
         \centering
         \includegraphics[width=\textwidth]{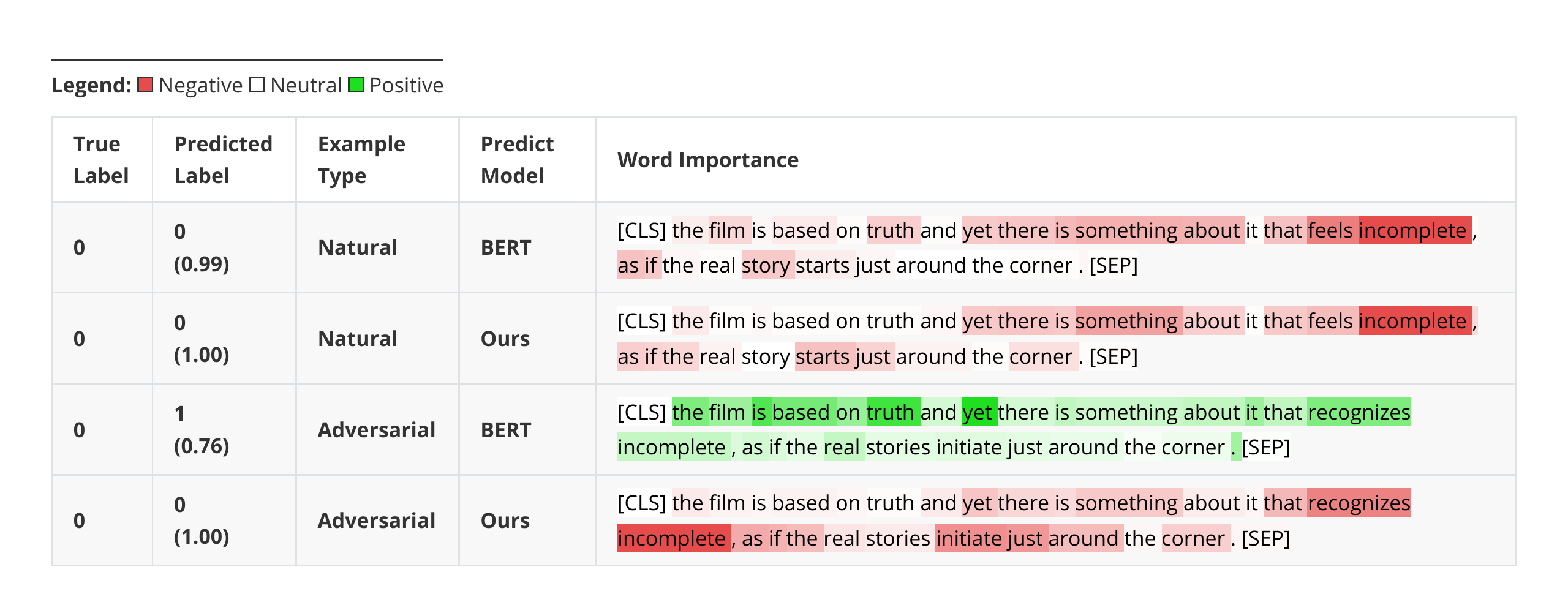}
         \caption{Input text: \emph{the film is based on truth and yet there is something about it that feels incomplete , as if the real story starts just around the corner .} }
         \label{fig:case2}
     \end{subfigure}
     \hfill
     \begin{subfigure}[b]{\textwidth}
         \centering
         \includegraphics[width=\textwidth]{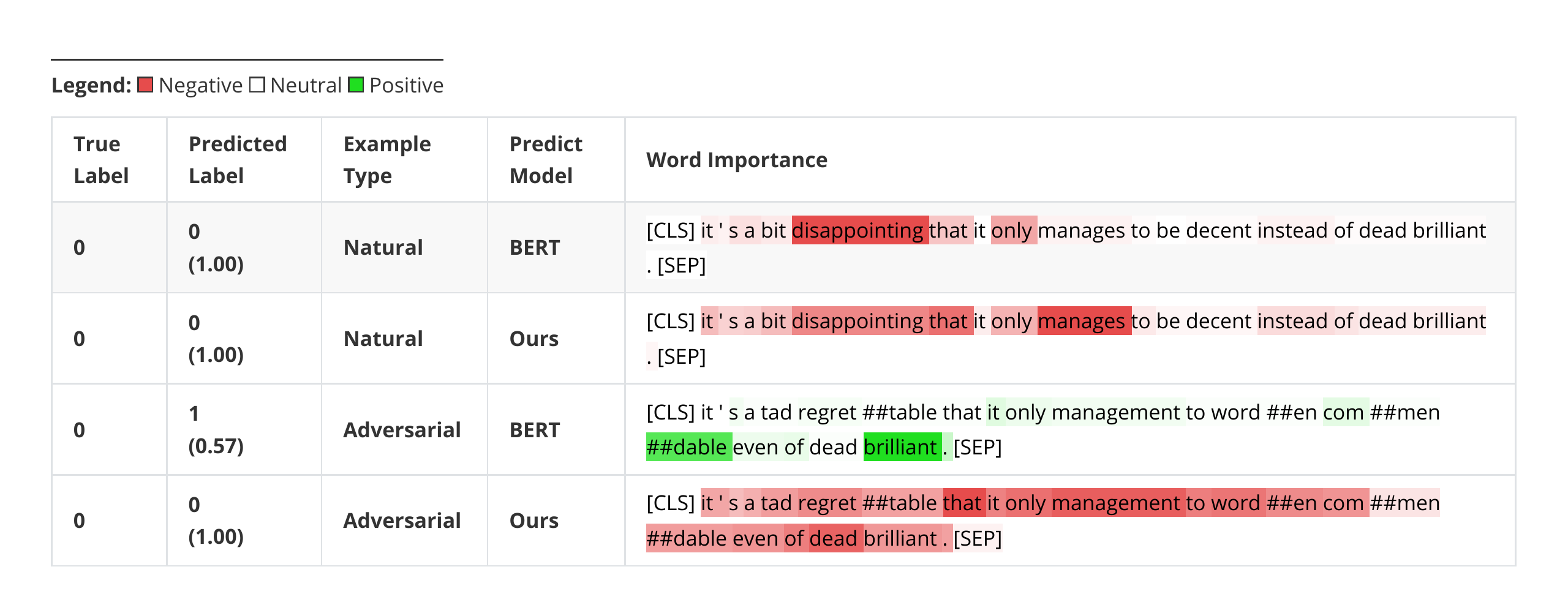}
         \caption{Input text: \emph{it 's a bit disappointing that it only manages to be decent instead of dead brilliant .}}
         \label{fig:case3}
     \end{subfigure}
        \caption{Case study examples. Each input text is tokenized and appended with special tokens, and \text{\#\#} is the tokenization mark used in BERT tokenizer. The darker the color, the more important the word is.}
        \label{fig:case}
\end{figure*}

To investigate whether our method helps the model to rely on deeper language phenomena for adversarial robustness, we visualize some examples where BERT is fooled under the attack while our method is robust. Figure \ref{fig:case} shows three examples selected from SST-2 dataset and the corresponding adversarial examples under TextFooler attack. We visualize the contribution of each token using a layer-wise relevance propagation based method \footnote{\url{https://github.com/hila-chefer/Transformer-Explainability}}, which integrates relevancy and gradient information through attention graph in Transformer. The darker the color, the greater contribution the word has to the prediction result.

We observe that BERT fine-tuning tends to rely on the most important tokens rather than the linguistic meaning of the sentence to make prediction. For instance, in figure \ref{fig:case1}, the prediction result of BERT mostly relies on the word \emph{hopeless}, and thus fails to make correct prediction when it is replaced. In contrast, in our method, most tokens contribute to the prediction, making it more robust to single word replacement.

\section{License}
The attack toolkit \emph{TextAttack} uses MIT License. The BERT is provided by \emph{huggingface transformers} which use Apache License 2.0.

\end{document}